%% file: root.tex
\documentclass[letterpaper, 10 pt, conference]{ieeeconf}  

\IEEEoverridecommandlockouts                              

\usepackage{subfiles}

\usepackage[utf8]{inputenc}

\usepackage{graphics}
\usepackage{hyphenat}
\usepackage{tikz}
\usepackage{amsmath}
\usepackage{amssymb}
\usepackage{array}
\usepackage{balance}

\usepackage{caption,tabularx,booktabs}
\usepackage{pifont}
\usepackage{algorithm}

\usepackage{algpseudocode}
\algnewcommand\Input{\item[\textbf{Input:}]}
   \algnewcommand\Output{\item[\textbf{Output:}]}

\usepackage{xcolor,colortbl}
\usepackage{multirow}
\usepackage{wrapfig}

\newcommand{\technique}{\textit{ASGARD}}

\newcommand{\attack}[1]{\emph{action-space attacks}}
\newcommand{\tcorrectaction}[1]{$\tilde{\bar{a_{t}}}$}
\newcommand{\scorrectaction}[1]{$\tilde{a_{t}}$}
\newcommand{\tcorruptaction}[1]{$\bar{a}\acute{}_t$}
\newcommand{\scorruptaction}[1]{$a\acute{}_t$}

\usepackage{listings}
\usepackage{makecell}
\usepackage{arydshln}
\usepackage{enumerate}
\usepackage{diagbox}
\usepackage{rotating}
\usepackage{subcaption}
\usepackage{float}
\usepackage{bbding}
\usepackage{mathtools}

\newlength{\imgdrop}
\definecolor{commentsColor}{rgb}{0, 0.5, 0}
\definecolor{keywordsColor}{rgb}{0.000000, 0.000000, 0.635294}
\definecolor{stringColor}{rgb}{0.558215, 0.000000, 0.135316}
\definecolor{applegreen}{rgb}{0.55, 0.71, 0.0}
\lstdefinestyle{cppstyle}{
    language=C++,
    backgroundcolor=\color{gray!10},
    basicstyle=\ttfamily\footnotesize,
    commentstyle=\itshape\color{gray},
    keywordstyle=\color{blue},
    stringstyle=\color{red},
    numbers=left,
    numberstyle=\tiny\ttfamily,
    numbersep=5pt,
    frame=lines,
    framerule=0.4pt,
    breaklines=true,
    tabsize=4,
    showstringspaces=false,
    xleftmargin=10pt,
    xrightmargin=0pt,
    belowskip=0pt,
    aboveskip=3pt,
    lineskip=-0.5pt,
    columns=flexible,
}

\lstdefinelanguage{myLang}
{
	morekeywords={bool, u32_int},
	sensitive=false,
	morecomment=[l]{//},
	morecomment=[s]{/*}{*/},
	morestring=[b]"
}
\lstdefinelanguage{myLangLL}
{
	morekeywords={entry, alloca, \define, i32, i64, call, void},
	sensitive=false,
	morecomment=[l]{//},
	morecomment=[s]{/*}{*/},
	morestring=[b]"
}

\title{\LARGE \bf
\technique{}: Action-Space Guard for UAV Resilience via \\ Reinforcement Learning
}

\author{
{\rm Mohsen Salehi}\\
The University of British Columbia\\
Vancouver, Canada\\
msalehi@ece.ubc.ca
\and
{\rm Karthik Pattabiraman}\\
The University of British Columbia\\
Vancouver, Canada\\
karthikp@ece.ubc.ca
} 

\begin{document}

\maketitle
\thispagestyle{empty}
\pagestyle{empty}

\begin{abstract}
Reinforcement learning (RL) controllers have been recently adopted for Unmanned Aerial Vehicles (UAV) navigation and control. However, they are susceptible to \attack{} that overwrite the action commands 
after the policy generates them and before the actuators execute them. 
While most existing defenses target attacks on the policy's inputs, those addressing \attack{} retrain the policy at training time and are not resilient to corrupted actions at runtime.
We propose \technique{}, a two-phase teacher--student pipeline for making RL-based UAV control resilient to \attack{}. 
In the teacher phase, an encoder combines the UAV's physical state with action-attack-related privileged information 
to produce an \emph{action-attack-aware} latent that trains the RL control policy and a monitor that outputs corrected action commands to the actuators.
In the student phase, both the encoder and the monitor are trained via supervised learning from their teacher counterparts 
to run on-board using only the UAV's physical state history.
We evaluate \technique{} across attack scenarios targeting different action commands on UAV. 
We find that \technique{} is resilient to \attack{} and completes the missions despite the attack. 
We further find that \technique{} generalizes to unseen attacks and remains resilient against stealthy attacks.
\end{abstract}


\input{Sections/Intro}

\input{Sections/Background}
\label{backgroundsec}

\input{Sections/Related}
\label{releatedsec}

\input{Sections/Design}
\label{designsec}

\input{Sections/Evaluation}
\label{evaluatesec}


\input{Sections/Conclusion}
\label{conclusionsec}










\bibliographystyle{IEEEtran}
\bibliography{references}

\end{document}

%% file: Sections/Intro.tex
\section{Introduction}
\label{sec-intro}

Unmanned Aerial Vehicles (UAVs) 
 consist of control software and hardware that rely on onboard sensors to observe the vehicle's physical state and on actuators to execute the commands that drive autonomous navigation.
The safety and correctness of this entire pipeline, from sensing and control software to actuation, are critical to mission success, 
yet UAVs remain vulnerable to a range of attacks~\cite{ilahi2021challenges}.
In particular, 
\emph{physical attacks} such as GPS spoofing that inject noise into the physical channel~\cite{humphreys2008assessing,dash2025armor},
and \attack{} that modify the \emph{control action commands}~\cite{lee2020spatiotemporally,alemzadeh2016targeted} (e.g., roll and pitch), threaten mission safety. 

In recent years, model-free reinforcement learning (RL) has been widely adopted across the UAV stack, from low-level control to autonomous navigation, because it adapts well to diverse and complex environments~\cite{chen2020learning,lee2020learning,hwangbo2017control}.
Building on this line of work, robust-RL methods such as ARMOR~\cite{dash2025armor} and robust adversarial reinforcement learning (RARL)~\cite{pinto2017robust} train control policies that remain reliable under sensor spoofing, offering a defense against \emph{physical attacks}.

While physical attacks have been addressed by existing techniques,
attackers can still 
modify \emph{control action commands} after the control policy has generated the commands - these are known as  \attack{}~\cite{lee2020spatiotemporally}.
 For example, modifying a single action command such as \emph{roll} or \emph{pitch} after inference can cause crashes or drive a UAV off its planned trajectory. 
The attacker can achieve this through different means, such as \textit{data-only attacks}~\cite{chen2005non,szekeres2013sok,alemzadeh2016targeted} (a type of software-based memory corruption attack), which modify the action commands \emph{after} the control policy has generated them. 
As a result, the control policy keeps producing action commands that look safe and correct on the surface, but the vehicle sees the corrupted actions and performs them.
Because these attacks occur \emph{after} the control policy has generated the actions, they cannot be detected by input-side resilient learning policies such as ARMOR and RARL.
Defenses that target \attack{} instead harden the policy against action perturbations encountered during training~\cite{tessler2019action,tan2020robustifying,lee2021query}; 
once deployed, they forward every action the policy produces directly to the actuators, with no mechanism to intervene if that action has been corrupted.

To close this gap, we propose \technique{}\footnote{\technique{} stands for Action-Space GuARD, and is named after the walled realm of the Norse gods.}, an RL-based control pipeline that makes UAV controllers resilient to \attack{}.
\technique{} is built on three innovations. 

First, inspired by prior work~\cite{lee2020learning,dash2025armor}, \technique{} uses a two-phase \emph{teacher}--\emph{student} training scheme.
\technique{} first uses a \emph{teacher encoder} with privileged information about \attack{}, 
such as the target action and attack duration, to generate an \emph{action-attack-aware} latent, which is then used to train the \emph{RL controller}.
The teacher encoder then supervises a \emph{student encoder} that runs on the device at inference time, 
generating a matching latent from only the UAV's physical state, so that the RL controller can operate without privileged information at deployment.

Second, \technique{} places a lightweight multilayer perceptron (MLP), called \emph{monitor}, between the RL controller and the actuators.
The monitor is trained under the same teacher--student scheme as the RL controller: a \emph{teacher monitor} is trained on the teacher encoder's output, and a \emph{student monitor} is trained on the student encoder's output under supervision from the teacher monitor.
Its lightweight design keeps the pipeline within the real-time constraints of onboard UAV control. 
Further, its placement shortens the attacker's effective tampering window and introduces a trust boundary between the RL policy and the actuators.

Third, unlike detection-only defenses that halt or land when they suspect an attack, \technique{}'s monitor actively corrects the RL controller's outputs at every step and forwards safe action commands to the actuators, so the UAV keeps flying under a safe control signal rather than falling back to a degraded mode or hovering and finally crashing.


\technique{} has three advantages over conventional techniques. 
First, because it is trained end-to-end on both clean and attacked timesteps, the monitor learns to pass safe action commands through unchanged and repair modified ones before they reach the actuators, without a separate detection step. 
Furthermore, since the teacher already sees the attack context through its privileged inputs, \technique{} does not need to keep generating fresh attacks during training, which keeps training costs low.
Finally, a trained \technique{} model can handle even attacks it did not see during training, without requiring re-training
, thereby making it robust to new attacks.



\textbf{Contributions.} We make three contributions as follows.

\begin{itemize}
    \item We provide resilience against \attack{} on RL-based UAV controllers by recovering action commands that are corrupted at runtime after they are generated by the policy, unlike existing action-space defenses, which retrain the policy against perturbations at training time instead of recovering corrupted actions.
    \item We propose \technique{}, a two-phase \emph{teacher}--\emph{student} control pipeline in which a teacher based on a Variational Autoencoder (VAE) with access to action-attack-aware privileged information supervises a runtime student, implemented as a Long Short-Term Memory (LSTM) network, that adapts the trained RL controller to rely only on the history of UAV physical states. 
    \item We introduce the monitor, a small multilayer perceptron (MLP) placed between the RL controller and the actuators that actively corrects each outgoing action command rather than merely detecting attacks. The monitor is trained under the same teacher--student scheme: a teacher monitor conditioned on the teacher's attack-aware latent supervises a student monitor that relies on the student latent at deployment, letting safe commands through unchanged and repairing corrupted ones before they reach the actuators. 
\end{itemize}
The results show that \technique{} is resilient against action-space attacks, completing 95\% of missions with no crashes when a single action command is corrupted, 
where an RL-only controller and ARMOR (the state-of-the-art resilient RL controller) complete 40\% and 50\% respectively and crash in 47\% to 36\% of missions. 
When all four commands are corrupted at once and both prior techniques fail every mission, \technique{} completes 67\%. 
\technique{} further generalizes to remain resilient to attack channels unseen during training and maintains resilience against stealthy attack patterns.

%% file: Sections/Background.tex
\section{Background}

\subsection{UAV Controller}
\label{sec:bg-uav-controller}

UAVs consist of control software (controller) and hardware, including sensors such as GPS and IMU that observe the environment, and actuators (the motors) that execute the action commands (e.g., roll and pitch channels) generated by the controller to complete a mission.
At each control step $t$, an onboard estimator fuses these sensor measurements into a physical state $s_t$ (e.g., position $p_t = (x,y,z)$ and velocity $v_t = (\dot{x},\dot{y},\dot{z})$).
Given $s_t$ and a reference $g_t$ drawn from the mission's waypoint list, the control policy $\pi$ produces the action commands $a_t = \pi(s_t, g_t)$, which are sent to the actuators. 
The action command is a tuple of four values: pitch $a_t^{p}$, roll $a_t^{r}$, thrust $a_t^{T}$, and gain $a_t^{K}$.
The four channels together parameterize the vehicle's intended motion: $a_t^{p}$ commands the forward/backward movement, $a_t^{r}$ the lateral movement, $a_t^{T}$ the vertical movement, and $a_t^{K}$ scales the overall speed at which the UAV moves toward its destination.

A mission is safe and successful when the UAV reaches every waypoint while staying within a mission-specific safety radius $\epsilon$ of the reference, 
i.e., when the tracking error $\Delta p_t = \|p_t - g_t\| \le \epsilon$ for all $t$. 

\subsection{Action-Space Attacks}
\label{sec:bg-doa}

\emph{Action-space attacks} intercept the RL controller's action commands after the control policy has produced them, and modify the values before they reach the actuators. 
To carry out this attack, attackers can use existing techniques such as memory corruption exploits (e.g., \emph{data-only attacks})~\cite{chen2005non,alemzadeh2016targeted}. 
In a UAV controller, the action commands $a_t = \pi(s_t, g_t)$ live in memory between the moment the control policy writes them and the moment the actuator reads them, so an attacker who can overwrite them replaces $a_t$ with modified commands $\tilde{a}_t$ (e.g., an additive bias $\tilde{a}_t = a_t + b_t$ chosen by the attacker) during this window.
Thus, the control policy and any check performed on its inputs still appear normal, and the unsafe behavior appears only in the physical world.
The damage depends on which action channel is targeted.
For instance, a perturbation of $a_t^{p}$ or $a_t^{r}$ causes unintended forward or sideways motion, while tampering with $a_t^{T}$ or $a_t^{K}$ drives it up, down, or past the destination; in either case, the UAV eventually deviates from the planned trajectory or crashes.
Hardware-level safety interlocks, if present, are designed to catch catastrophic failures (e.g., complete motor loss) and are therefore ineffective against the subtle, gradual action corruptions we consider.

\subsection{Threat Model}
\label{sec:bg-threat}

We consider an adversary that can overwrite the action commands $a_t$ after the control policy $\pi$ has produced them and before the actuators execute the actions.
For instance, at each control step, the adversary may replace $a_t$ with modified commands $\tilde{a}_t = a_t + b_t$, where $b_t$ is an attacker-chosen bias applied to one or more of the four action channels (pitch $a_t^{p}$, roll $a_t^{r}$, thrust $a_t^{T}$, and gain $a_t^{K}$) with varying severity, patterns, and duration.
We assume the attacker can act stealthily by applying small biases $b_t$ that cause gradual drift from the intended trajectory.
The encoder, the control policy, and the monitor are immutable after training, and hence we assume they reside in read-only memory at deployment and cannot be modified by the adversary.
Action commands, in contrast, are recomputed at every control step and must hence be stored in writable memory, leaving them exposed to corruption by attackers.

Physical attacks on sensors, such as GPS spoofing, are outside our scope; they target a different point in the pipeline and are addressed by prior work~\cite{dash2025armor,pinto2017robust} 
Attacks on the ground station, the mission plan, or the communication link between them are likewise out of scope.

%% file: Sections/Related.tex
\section{Related work}

%
We classify related work into two broad categories. 
\textbf{Model-free Reinforcement Learning}
Recent advances in reinforcement learning (RL) have made learned policies a leading approach for robotic control, replacing hand-tuned controllers with end-to-end training on interaction data.
Applications range from low-level quadrotor stabilization~\cite{hwangbo2017control} and championship-level drone racing~\cite{kaufmann2023champion} to legged locomotion over challenging terrain~\cite{lee2020learning}.
Extending this line of work, other approaches adapt RL to robust and safe operation: co-training the policy against 
a perturbing adversary~\cite{pinto2017robust}, learning under environmental uncertainty~\cite{fan2020learning,sacerdoti2024reinforcement} 
to adapt the robot's behavior in unseen environments,
or keeping the system in a safe state using manually defined, fixed boundaries of unsafe actions~\cite{cheng2019end}, 
which require anticipating unsafe regions in advance and are therefore not designed for action-space attacks that occur after the action is generated 
by the control policy, the threat model considered by \technique{}.

\textbf{Attacks on Robots}
Adversarial attacks on RL-based techniques span several categories~\cite{ilahi2021challenges,huang2017adversarial}, including perturbations to observed states, manipulations of the training environment, and attacks on the action space.
Due to space constraints, we focus on the two most relevant to this work.

\noindent
\textit{1. Physical Attacks} perturb the sensor measurements that feed the policy, such as GPS spoofing~\cite{humphreys2008assessing}.
Robust Adversarial RL (RARL)~\cite{pinto2017robust} co-trains a policy against an adversary that injects adversarial perturbations as external forces on the agent's body, modeling physical-layer disturbances such as mass or friction mismatch;
ARMOR~\cite{dash2025armor} trains a UAV controller under a teacher--student scheme with privileged information about sensor spoofing. 

\noindent
\textit{2. Action-Space Attacks}
target the action space of an RL control policy.
Lee et al.~\cite{lee2020spatiotemporally} explore action-space adversarial attacks under spatial and temporal budgets, 
comparing an attack that perturbs actions independently at each step against one that plans across multiple steps using the agent's dynamics. 
They show these attacks are effective at degrading a Deep RL (DRL) agent's performance in simulated environments. 
However, their work focuses purely on crafting stronger attacks, with no defense or resilience mechanism proposed.
Several papers have proposed robustifying DRL agents against action-space perturbations by training the policy against adversarial or worst-case perturbations to its actions~\cite{tessler2019action,tan2020robustifying}.
Similarly, Lee et al.~\cite{lee2021query} investigate black-box targeted attacks using a learned adversarial policy and show that fine-tuning the nominal 
policy via adversarial training can partially mitigate such attacks, though it does not fully eliminate their success. 
However, these approaches retrain the policy against such attacks at training time rather than addressing corrupted actions at runtime after they are generated by the policy, which is the problem \technique{} addresses.
Different attack techniques can be used to corrupt the RL controller's action commands, 
including \emph{non-control-data attacks}~\cite{chen2005non}, software-based exploits that manipulate program data (e.g., action commands in UAVs).

%% file: Sections/Design.tex
\section{Design: \technique{}}
\label{sec-design}
We first present an overview of the design of \technique, followed by a deep dive into the teacher and student phases. 
\subsection{\technique{}: Overview}

\begin{figure*}[ht]    
	\centering
	\includegraphics[scale=0.63]{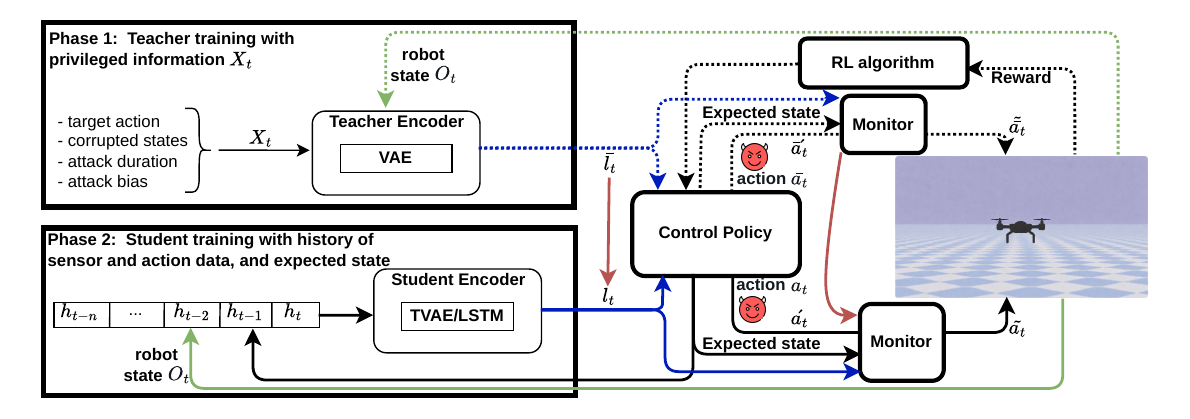} 
	\caption{Overview of \technique{}'s operation.
	\textbf{(1) Teacher Phase:} the Teacher encoder combines privileged information with the UAV's physical state to produce an action-attack-aware latent that trains the RL control policy and the Teacher monitor, which outputs corrected action commands to the actuators.
	\textbf{(2) Student Phase:} the Student encoder and Student monitor are trained via supervised learning from their Teacher counterparts, using only the UAV's physical state history as input; these three components (Student Encoder, Control Policy, Student monitor) run together on-board at deployment.}
	\label{img:Overview}
\end{figure*}

Figure~\ref{img:Overview} shows an overview of \technique{}, a two-phase teacher--student pipeline with three components in each phase for defending against \attack{}.
In the teacher phase, the teacher encoder consumes the UAV's physical state ($o_{t}$) along with action-attack-related privileged information ($x_{t}$) (e.g., target action) to produce an \emph{action-attack-aware latent} ($\bar{l_{t}}$).
This latent trains the RL control policy, which in turn outputs both an action ($\bar{a_{t}}$) and an expected next state, 
jointly training the teacher monitor on the latent, the action, and the expected state to output a corrected action (\tcorrectaction{}).
The three components are trained end-to-end on a mix of clean and attacked trajectories, so the monitor learns both to forward safe commands and to repair corrupted ones (\tcorruptaction{}). 

In the student phase, the pipeline is adapted for onboard deployment, where privileged information is unavailable.
The student encoder is trained via supervised learning from the teacher encoder to match its attack-aware latent from only the UAV's physical state history (H),
and the student monitor is trained under supervision from the teacher monitor using the student latent ($l_{t}$), 
the policy's expected state, and the generated action commands.
The control policy is carried over from the teacher phase and now consumes the student latent to produce action commands ($a_{t}$).
At runtime, the student encoder, control policy, and student monitor execute together on the onboard device; at each step, 
the monitor forwards the correct action commands to the actuators (\scorrectaction{}).

\subsection{\technique{}: Teacher Phase}
Like prior work~\cite{lee2020learning,aljalbout2024role}, we formulate the control problem as a Markov Decision Process (MDP), 
defined by the tuple $(\mathcal{S}, \mathcal{A}, \mathcal{P}, r)$, corresponding respectively to the state space, action space, 
transition probability $\mathcal{P} : \mathcal{S} \times \mathcal{A} \to \mathcal{S}$, and scalar reward function $r$. 
Training proceeds by selecting an action $a_t$ from a control policy $\pi(a_t \mid s_t)$, receiving a reward $r_t$, 
and maximizing the expected discounted sum of rewards over time. 

In the first phase of training pipeline, \technique{} teacher assumes access to both the UAV's physical state $o_t$ 
including the UAV's position, orientation, linear velocity, and angular velocity ($o_t = \left[x, y, z, \phi, \theta, \psi, \dot{x}, \dot{y}, \dot{z}, \dot{\phi}, \dot{\theta}, \dot{\psi}\right]$)
and privileged information $x_t$ describing the \attack{} information and its corrupted physical states,
the corruption bias currently applied, and its duration. 
The teacher's full state is thus $s_t \coloneqq \langle o_t, x_t \rangle$. 
The teacher encoder $E_{\text{teacher}}$, implemented as a variational autoencoder (VAE)~\cite{kingma2013auto},
maps $s_t$ ($E_{\text{teacher}}(o_t, x_t)$) to an \emph{action-attack-aware} latent representation $\bar{l}_t$,
the input reconstruction $\hat{s_{t}}$ and
the prediction of type of attack $\hat{y_{t}}$. 

The control policy $\pi(a_t \mid \bar{l}_t)$ is trained via Proximal Policy Optimization (PPO)~\cite{schulman2017proximal}, 
using $\bar{l}_t$ as input to produce the intended action $\bar{a}_t$. 
For estimating and correcting action commands, alongside the policy, 
we train an auxiliary \emph{expected-state head} $F$ ($\hat{s}_{t+1} = F(\bar{l}_t)$), which predicts the physical state the UAV should reach immediately following $\bar{a}_t$.
The expected state will help the monitor in the next step to correct or unchange the received action commands based on the current latent. 

$F$ is trained via supervised regression against the UAV's true next state, observed one step ahead under \emph{clean, unattacked} dynamics since policy produces action commands, it has access to uncorrupted ones:
\begin{equation}
    \mathcal{L}_{\text{state}} = \left\| F(\bar{l}_t) - s^{\text{clean}}_{t+1} \right\|^2
    \label{eq:loss-state}
\end{equation}
computed on uncorrupted commands, so that $\hat{s}_{t+1}$ always reflects the physically correct outcome, independent of whether an attack occurs at $t$.

The teacher monitor ($M_{\text{teacher}}$) is the core mechanism enabling \technique{} to recover from action-attacks. 
At every timestep, the attacker may overwrite the policy's intended action $\bar{a}_t$ with a corrupted action \tcorruptaction{}. 
$M_{\text{teacher}}$ observes the latent $\bar{l}_t$, the received (possibly corrupted) action \tcorruptaction{}, and the expected next state $\hat{s}_{t+1}$, and outputs a corrected action:
\begin{equation}
    \tilde{\bar{a_{t}}} = M_{\text{teacher}}(\bar{l}_t, \bar{a}\acute{}_t, \hat{s}_{t+1})
    \label{eq:teacher-monitor}
\end{equation}
$M_{\text{teacher}}$ is trained via supervised regression toward the uncorrupted action 
using a mixture of clean timesteps (where \tcorruptaction{} $ = \bar{a}_t$) and attacked timesteps (where \tcorruptaction{} $\neq \bar{a}_t$). 
Training on both populations with accessing to privileged action attack information helps 
$M_{\text{teacher}}$ to train on both scenarios and send commands consistent with the uncorrupted intent 
to the actuators. 

At every timestep of teacher-phase training, the executed action is \tcorrectaction{}, 
meaning $M_{\text{teacher}}$'s correction directly determines the UAV's physical trajectory, not merely a diagnostic signal computed alongside it. 
The reward $r_t$ that shapes the policy's training reflects progress toward the current target waypoint $g_t$, adopted by prior work:
\begin{equation}
	\label{eqn:reward-func}
	\begin{split}
	r_t = \; & \underbrace{R_{\text{goal}} \cdot \exp\left(-\lambda \|p_t - g_t\|\right)}_{\substack{\text{sharp reward as}\\\text{UAV nears goal}}} 
	- \underbrace{\alpha \|p_t - p_{t-1}\|}_{\substack{\text{penalize large}\\\text{position changes}}} \\
	& - \underbrace{\beta \theta_t}_{\substack{\text{penalize}\\\text{excessive tilt}}} 
	- \underbrace{\gamma \|a_t - a_{t-1}\|^2}_{\substack{\text{penalize abrupt}\\\text{control changes}}}
	\end{split}
\end{equation}
where $p_t$ is the UAV's position and penalty weights shown with $\alpha$ for deviation from target, $\beta$ for instability, $\gamma$ for sudden movement; 
upon reaching the final waypoint, a terminal bonus proportional to the remaining time steps in the episode rewards early, successful completion of the mission.

\subsection{\technique{}: Student Phase}

For onboard deployment, where privileged information $x_t$ is unavailable, 
we train a student encoder $E_{\text{student}}$ using only the UAV's recent physical state history $H_t := \{o_{t-N}, \dots, o_{t-1}\}$, 
together with the UAV's own previously executed actions $\{a_{t-N}, \dots, a_{t-1}\}$, 
allowing the student to account for the consequences of its own recent commands when estimating the current \emph{action-attack-aware} latent ($l_t = E_{\text{student}}(H_t)$).

Given that the underlying data forms a sequence of time-dependent measurements, $E_{\text{student}}$ is implemented as a 
temporal variational autoencoder (TVAE) using a
recurrent Long Short-Term Memory (LSTM) network, trained via supervised regression to approximate the teacher's latent.
\begin{equation}
    \mathcal{L}_{\text{feat}} = \left\| l_t - \bar{l}_t \right\|^2
    \label{eq:loss-feat}
\end{equation}
We additionally minimize the discrepancy between the actions the shared control policy produces from each latent:
\begin{equation}
    \mathcal{L}_{\text{act}} = \left\| \pi(l_t) - \pi(\bar{l}_t) \right\|^2
    \label{eq:loss-act}
\end{equation}
so that the student latent is optimized not only to numerically resemble $\bar{l}_t$, but to also yield the same downstream control decisions.
Overall, the student encoder tries to approximate the teacher encoder's output using supervised learning, and correspondingly 
produce similar actions generated by the policy, by minimizing the loss:
$\mathcal{L}_{\text{feat}} + \mathcal{L}_{\text{act}} + \mathcal{L}_{\text{attack}}$, where $\mathcal{L}_{\text{attack}}$ is the attack classification loss.

The student monitor $M_{\text{student}}$ shares $M_{\text{teacher}}$'s architecture.
Given the student latent $l_t$, the received action $a\acute{}_t$, and the student's own expected-state prediction, $M_{\text{student}}$ is trained to match the teacher monitor's correction for the same underlying scenario:
\begin{equation}
    \resizebox{\columnwidth}{!}{$
	\mathcal{L}_{\text{monitor}}^{\text{student}} = \left\| M_{\text{student}}(l_t, a\acute{}_t, \hat{s}^{\text{student}}_{t+1}) - M_{\text{teacher}}(\bar{l}_t, \bar{a}\acute{}_t, \hat{s}_{t+1}) \right\|^2    $}
    \label{eq:loss-monitor-student}
\end{equation}

where the same attack 
is applied to both the teacher's and student's action at each training step, 
ensuring the two monitors are supervised on matching scenarios. The control policy $\pi$ is reused unchanged from the teacher phase, 
since both phases share an identical action space and reward objective. 
At deployment, only $E_{\text{student}}$, $\pi$, and $M_{\text{student}}$ execute onboard: 
at each timestep, $M_{\text{student}}$ consumes the policy's action and either forwards it unchanged or replaces it with \scorrectaction{}
before it reaches the actuators.

In particular, at each control step, the student encoder computes $l_t$ from the sensor and action history; 
the control policy produces $a_t = \pi(l_t)$; and $M_{\text{student}}$ receives $(l_t, a\acute{}_t, \hat{s}_{t+1})$ unconditionally, on every step, 
regardless of whether an attack is present (\scorruptaction{} $ = a_t$ or \scorruptaction{} $\neq a_t$). 
$M_{\text{student}}$ outputs an action command that is used directly by the actuators, 
adapting toward the uncorrupted command the control policy would have produced, even if an attack occurs at that step.

%% file: Sections/Evaluation.tex
\section{\technique{} Evaluation}
This section first presents the experimental setup, and then  evaluates \technique{} along four aspects: 
(i) the training performance of the teacher--student pipeline (\S\ref{subsec:training-performance}), 
(ii) \technique{}'s behavior under action-space attacks compared with prior techniques (\S\ref{subsec:action-attacks}, \S\ref{subsec:comparison}), 
(iii) zero-shot performance against unseen attacks (\S\ref{subsec:zeroshot}), 
and (iv) resilience against stealthy attacks (\S\ref{subsec:stealthy}). 

\subsection{Experimental Setup}
For training and evaluation, we target a waypoint-following task where a quadcopter (shown on the right)
\begin{wrapfigure}[4]{r}{0.22\columnwidth}
    \raggedleft
    \raisebox{\dimexpr 0pt-\imgdrop\relax}[0pt][0pt]{%
        \includegraphics[width=0.19\columnwidth]{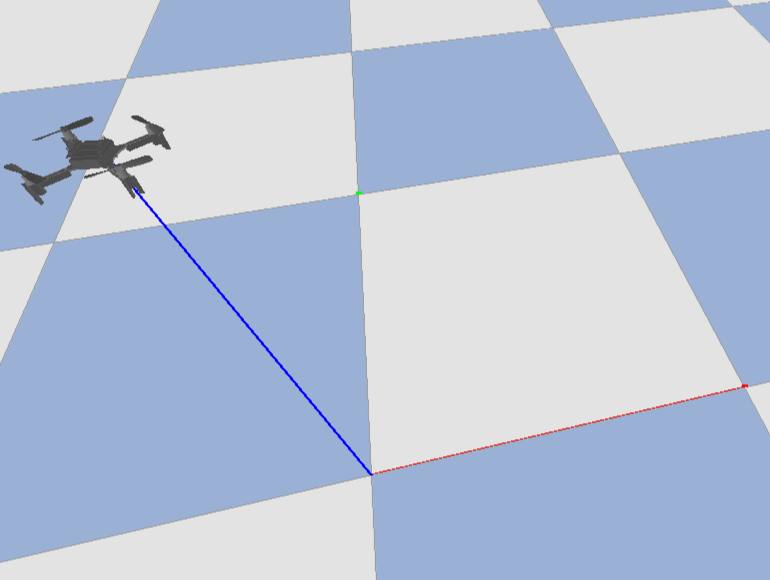}%
    }
\end{wrapfigure}
reaches randomly-selected 3D goal positions under coupled translational-rotational dynamics.
We build the environment on \emph{gym-pybullet}~\cite{panerati2021learning}, a PyBullet-based quadrotor simulator with an OpenAI Gym interface~\cite{brockman2016openai} and realistic rigid-body dynamics, matching the setup used in prior UAV RL work~\cite{yuan2022safe,dash2025armor}.

For \technique{}, we extend this environment with three custom components: an action-attack injection mechanism, a channel for privileged action-context observations, and a monitor correction interface.
We adopt ARMOR's publicly available implementation for the teacher encoder and control policy~\cite{dash2025armor}, modifying them to support our innovations in order to be resilient against \attack{}.
We implement the teacher--student encoder architecture in the training pipeline with a 15-dimensional latent, and the student encoder takes an 80-timestep history of the UAV's physical state as input. 
Further, the RL control policy is trained with PPO, and the monitors are implemented as lightweight MLPs.
Although \technique{} can be applied to other types of vehicles, we restrict our evaluation to a single quadrotor model due to space constraints, 
prioritizing depth of analysis on \attack{} over breadth across platforms.

Table~\ref{tab:attack-types} summarizes the five attack configurations we use to evaluate \technique{}'s resilience across a broad range of action-attack scenarios, each targeting a different action channel with its own bias range and duration.
A \emph{Pitch} attack disrupts the UAV's forward progress, at times reversing its direction of travel outright, 
while a \emph{Roll} attack turns what should be a small, controlled sideways adjustment into a hard, unintended lean. 
\emph{Thrust} attacks strike at the UAV's climb and descent rate, disrupting vertical stability over an extended period.
\emph{Gain} attacks act differently still, distorting not the direction of the UAV's movement but its strength, leaving it barely responsive one moment and lurching the next. 
Finally, we evaluate a combined attack that applies \emph{all four} simultaneously, representing a worst-case adversary that corrupts every command at once and drives the UAV off-course along all four control dimensions in parallel.


\begin{table}[h]
    \centering
    \resizebox{\columnwidth}{!}{%
    \begin{tabular}{c|c|c|c}
    \textbf{Target Action} & \textbf{Bias Range} & \textbf{Attack Duration} & \textbf{Explain}                             \\ \hline
    Pitch                  & (-4)-(4)            & max 90s                 & Disrupts or reverses forward progress         \\
    Roll                   & (-1)-(1)            & max 90s                   & Turns a small adjustment into a hard, unintended lean    \\
    Thrust               & (-5)-(5)            & max 120s                     & Disrupts climb/descent rate \\
    Gain                   & (-15)-(15)            & max 150s                   & Distorts movement strength           \\
    All                    & (-1)-(1)            & max 60s                   & All four simultaneously, worst-case scenario             \\ \hline
    \end{tabular}
    }
    \caption{Different types of attacks on each action command for evaluating \technique{}.}
    \label{tab:attack-types}
\end{table}

For comparison, we consider two additional techniques.
The state-of-the-art RL resilience approach \textbf{ARMOR}~\cite{dash2025armor} uses a similar teacher--student training pipeline but targets \emph{physical attacks} on the sensor inputs of the RL control policy, and forwards its action commands to the actuators without any correction step.
Other adversarial training defenses~\cite{fei2020learn,pinto2017robust} share the same idea, jointly training a control policy alongside an adversary that corrupts the observations fed to the control policy.
ARMOR has been shown to outperform such approaches, so we use it as the state-of-the-art baseline for our comparison.
Further, as an \textbf{Ablation Study} to show the effectiveness of \technique{}'s two-stage training pipeline, including the encoders and monitor,
we consider a \textit{Baseline-RL controller} that removes both from \technique{}'s architecture in two settings:
one with access to privileged action-attack information during training, and one without.
The results show that both settings perform comparably, confirming that training with privileged information alone, without the encoder
(no latent representation in either case) and monitor, is not sufficient for resilience against \attack{};
we therefore report both under the single label \textit{Baseline-RL}.

We adapt the prior work definition of metrics to evaluate \technique{} and its comparisons with prior work.
\textbf{Mission Success Rate} is the fraction of evaluated episodes in which the UAV reaches within its designated waypoint threshold ($\epsilon$ = 5m~\cite{dash2024specguard,dash2025armor}) before the episode ends ($\|dest - pos\| \le \epsilon$, where $dest$ is the target waypoint and $pos$ is the UAV's current position), without having triggered a crash or timeout termination first.
\textbf{State Drift} is the mean Euclidean distance, in meters, between $\text{pos}$ and $\text{dest}$, computed at each timestep during the attack and averaged across evaluation episodes.
\textbf{Crash Rate} is the fraction of evaluated episodes that terminate specifically because the UAV's state exceeds a predefined safety bound,
such as leaving the valid flight volume or tilting beyond a safe orientation, causing a crash.

For brevity, we use \emph{Teacher} and \emph{Student} to refer to \technique{}'s full pipeline (encoder, control policy, and monitor) at each phase.

\subsection{\technique{} Training Performance}
\label{subsec:training-performance}

\begin{figure}[h]
    \centering
    \begin{subfigure}[b]{0.48\linewidth}
        \centering
        \includegraphics[width=\linewidth]{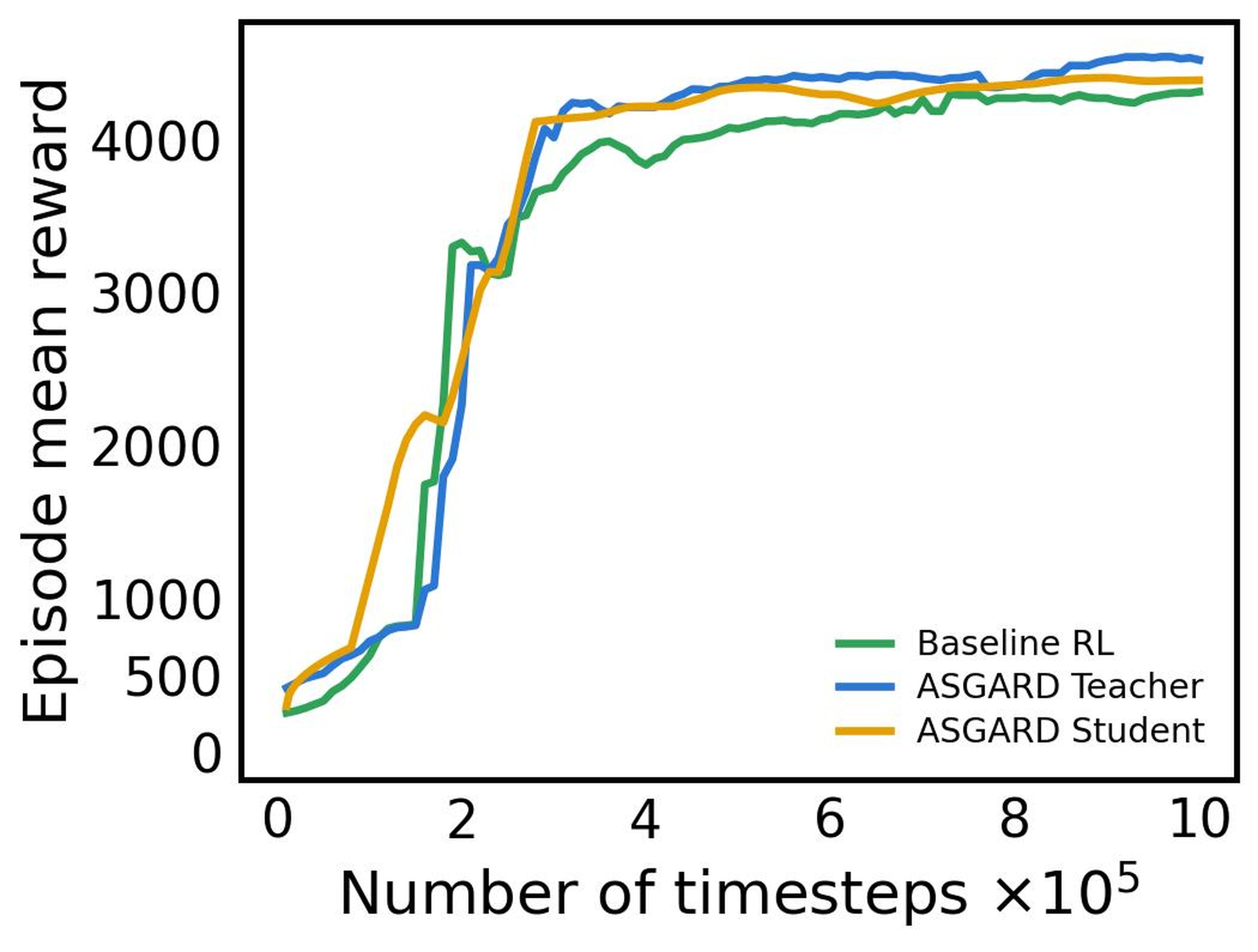}
    \end{subfigure}
    \hfill
    \begin{subfigure}[b]{0.48\linewidth}
        \centering
        \includegraphics[width=\linewidth]{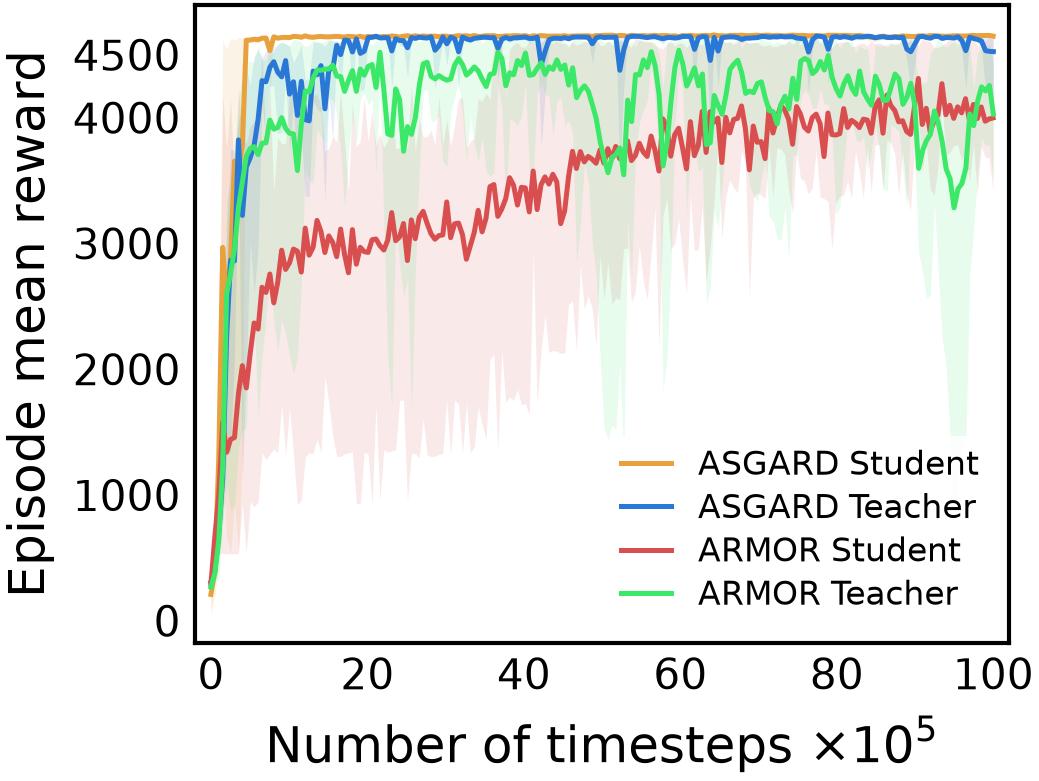}
    \end{subfigure}
    \caption{Training performance comparison. \textbf{Left (a):} Nominal conditions, all methods achieve similar final performance. 
    \textbf{Right (b):} Adversarial conditions with \attack{}, both \technique{} Teacher and Student converge faster than ARMOR while ARMOR has many fluctuations.}
    \label{fig:phoenix-training}
\end{figure}

To evaluate whether \technique{}'s Student can adapt to the Teacher's performance, we train the baseline-RL and \technique{}'s Teacher over $10 \times 10^5$ timesteps under nominal (attack-free) conditions, and then train \technique{}'s Student under the Teacher's supervision.
All training curves in Figure~\ref{fig:phoenix-training} are averaged over five runs with random seeds.
Figure~\ref{fig:phoenix-training}(a) shows the results. We make two observations.
First, \technique{}'s Teacher (blue) and Student (yellow) both reach the maximum episodic reward (approximately 4000) within about $3 \times 10^5$ timesteps, whereas the baseline-RL approach (green) requires roughly $7 \times 10^5$ timesteps to reach the same level.
Second, the Student closely tracks the Teacher throughout training and both reach the same maximum reward as the baseline-RL. This shows that the teacher--student encoding preserves learning performance under attack-free conditions.

To evaluate \technique{}'s effectiveness under \attack{}, we train both \technique{} and ARMOR under equivalent conditions, giving each teacher access to the privileged information and then supervising their respective students.
Figure~\ref{fig:phoenix-training}(b) shows the resulting training curves, with shaded regions indicating variation across the five runs. We can see that 
ARMOR's Teacher and Student both struggle to converge, reaching a maximum episodic reward of approximately 4000 after $100 \times 10^5$ timesteps and with large fluctuations across runs.
\technique{}, in contrast, converges to a higher episodic reward of approximately 4500 in under $20 \times 10^5$ timesteps ($5\times$ faster) with much tighter variance across runs.
Overall, these results show that \technique{} is effective under both nominal and adversarial conditions, outperforming both ARMOR and the baseline-RL, and that the Student reproduces the Teacher's action-attack resilience while relying on only the UAV's physical state history.
In the remaining subsections, we focus on the Student (the pipeline actually deployed on the device) and refer to it as \technique{}.




\subsection{\technique{} under Action Attacks}
\label{subsec:action-attacks}

To evaluate \technique{}'s effectiveness under \attack{}, we run the four single-channel attacks (Pitch, Roll, Thrust, Gain) and the combined \emph{All} attack from Table~\ref{tab:attack-types}.
As Table~\ref{tab:physical_attack} shows, \technique{} successfully completes an average of 95\% of missions across the four single-channel attacks without any crashes, while achieving the lowest state drift ($\sim$0.30m on average).
Furthermore, under the \emph{All} attack, which is also difficult for an attacker to carry out in practice, 
\technique{} successfully finishes 67\% of missions with only a 10\% crash rate while keeping the state drift small ($\sim$0.26m).

\begin{figure}[h]
    \centering
    \setlength{\tabcolsep}{1pt}      
    \renewcommand{\arraystretch}{1}
    \begin{tabular}{cccc}
    \includegraphics[width=0.24\linewidth]{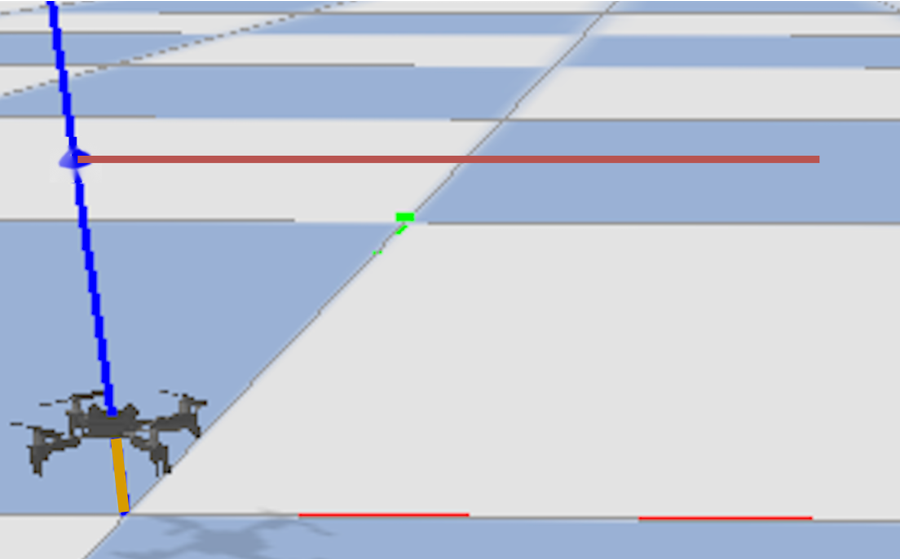} &
    \includegraphics[width=0.24\linewidth]{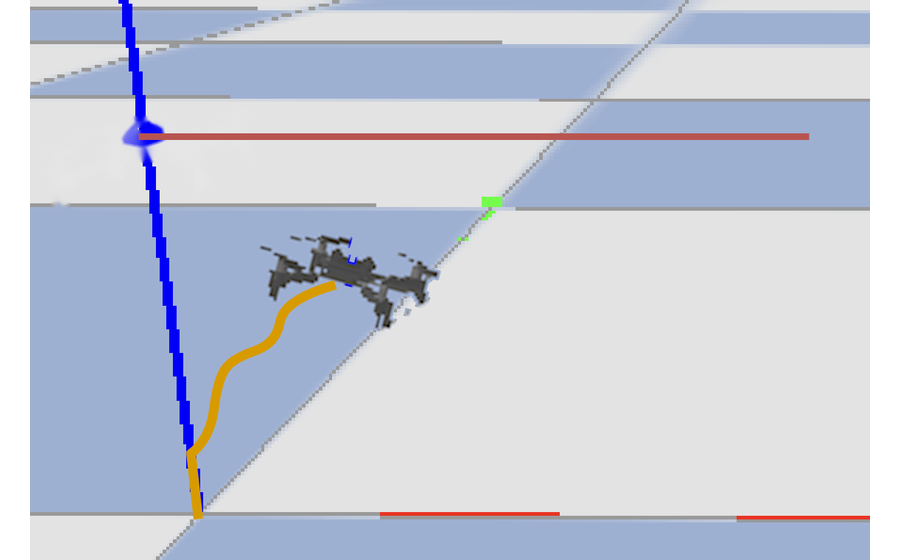} &
    \includegraphics[width=0.24\linewidth]{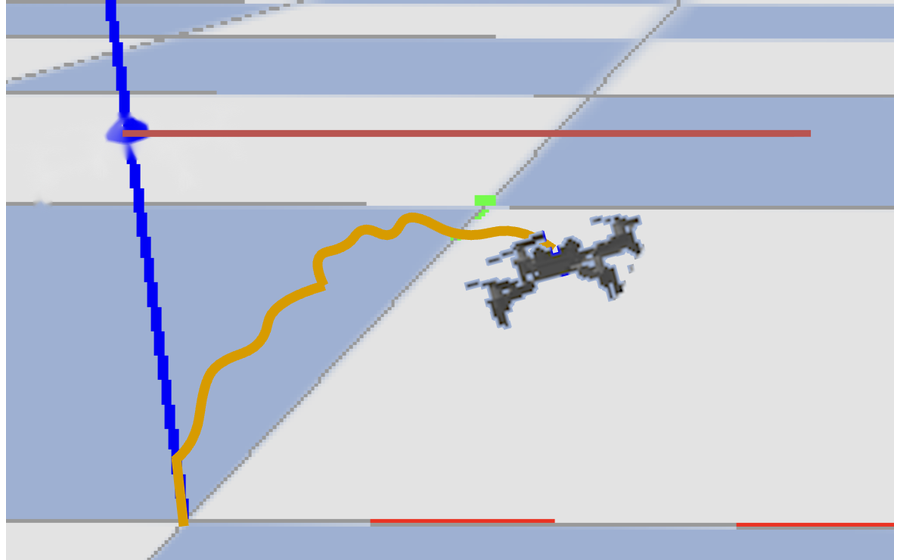} &
    \includegraphics[width=0.24\linewidth]{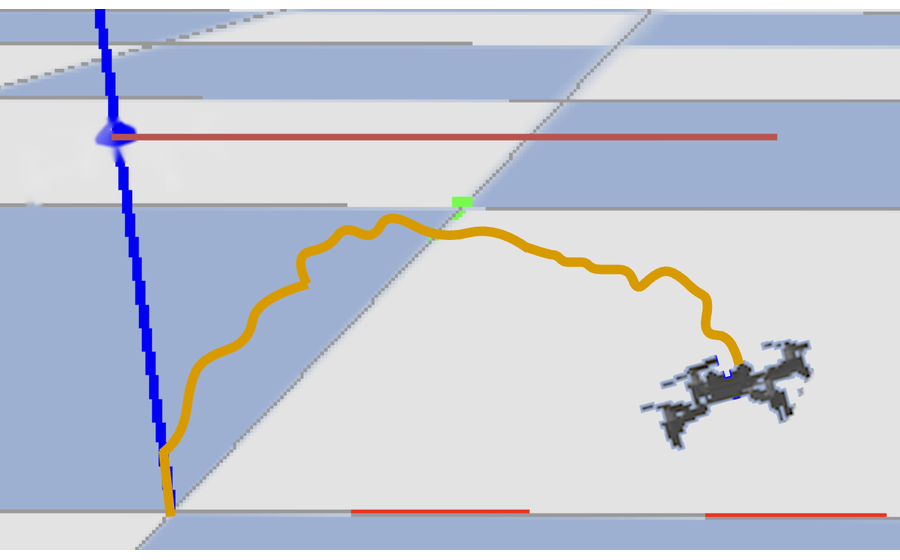} \\[1pt]
    
    \includegraphics[width=0.24\linewidth]{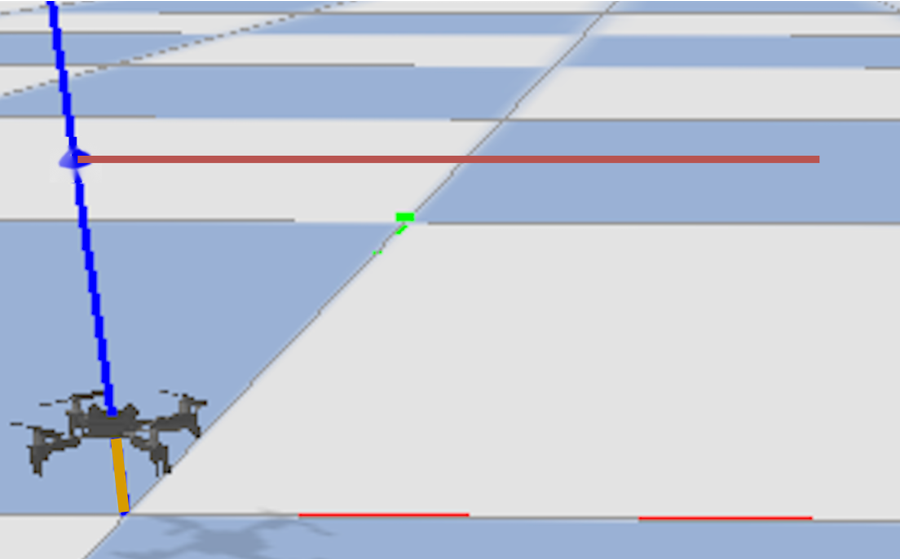} &
    \includegraphics[width=0.24\linewidth]{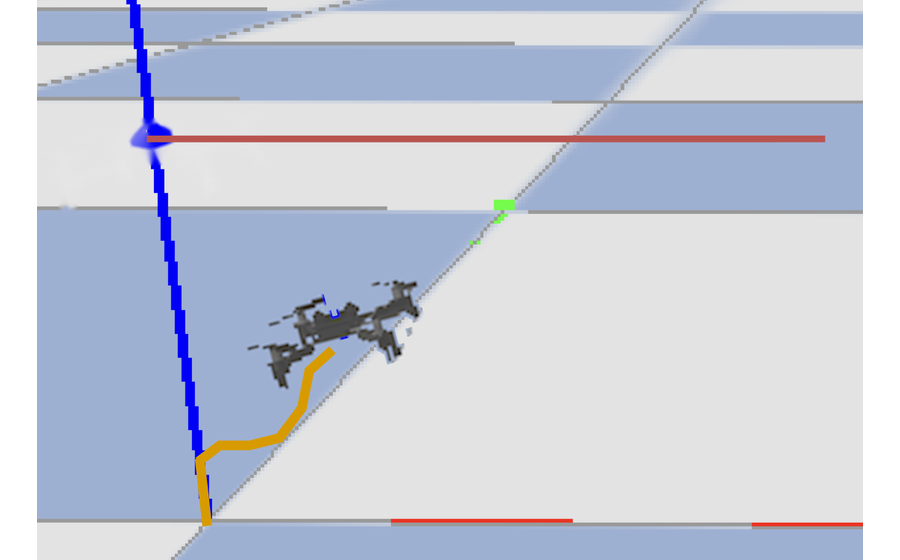} &
    \includegraphics[width=0.24\linewidth]{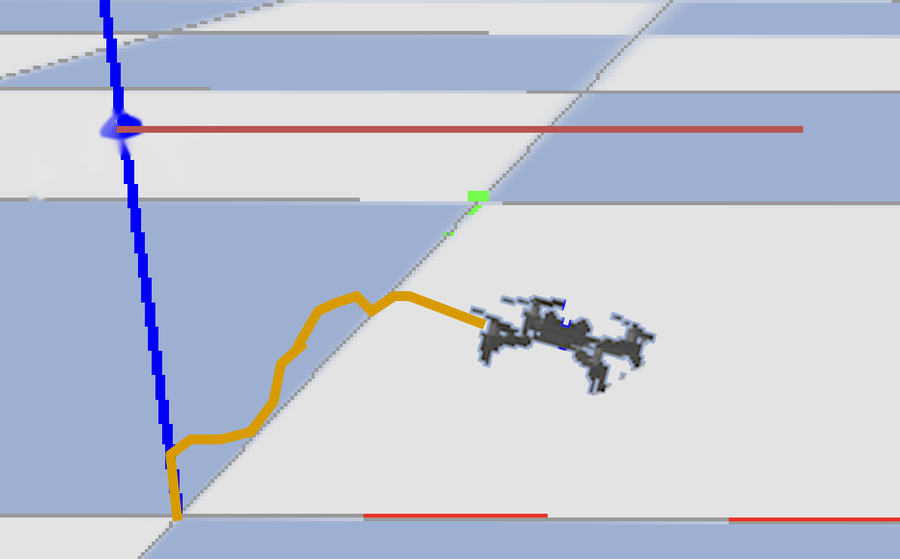} &
    \includegraphics[width=0.24\linewidth]{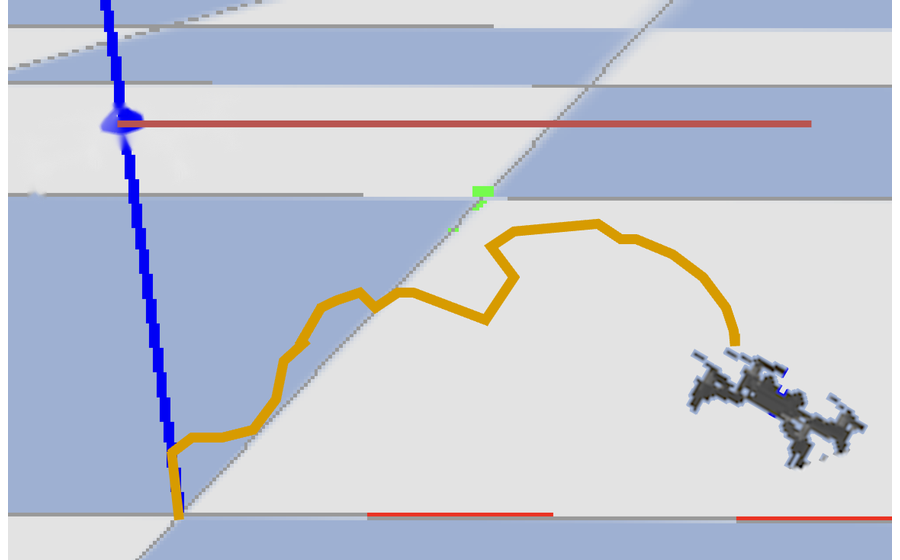} \\[1pt]
    
    \includegraphics[width=0.24\linewidth]{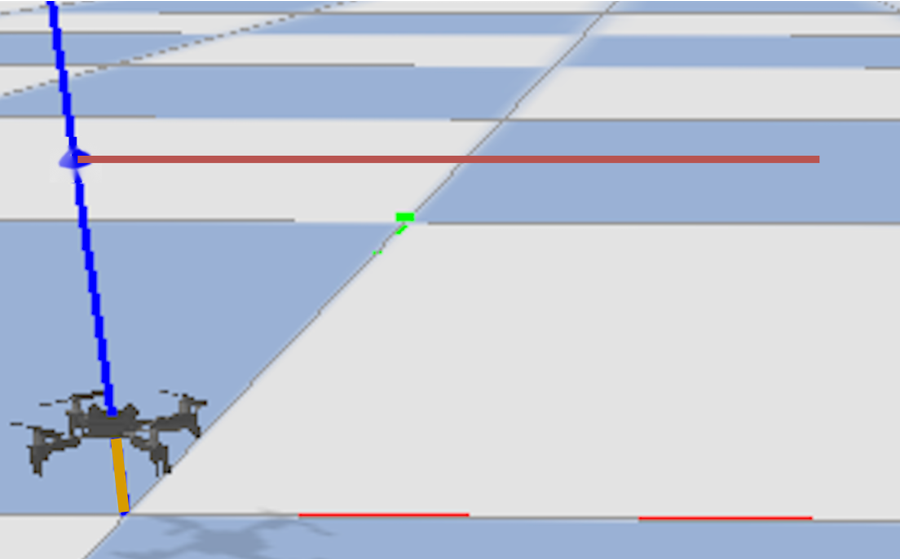} &
    \includegraphics[width=0.24\linewidth]{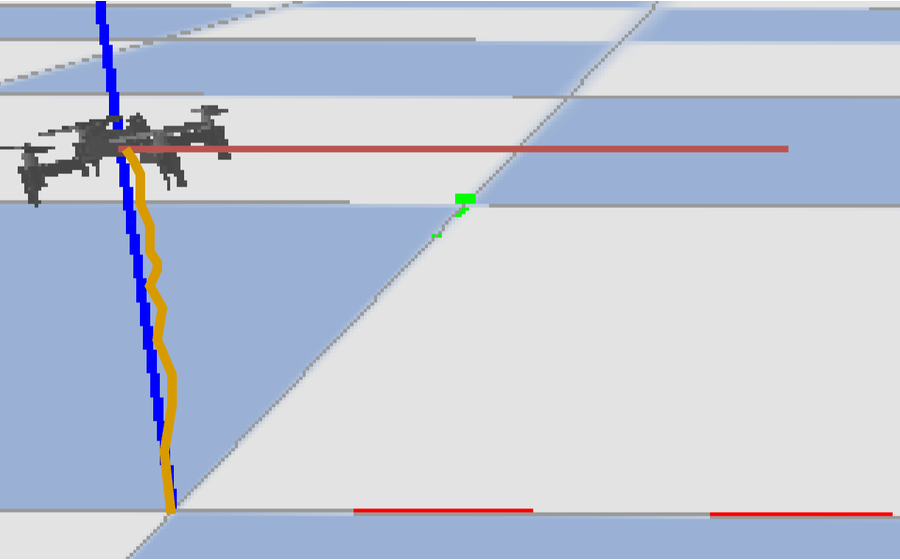} &
    \includegraphics[width=0.24\linewidth]{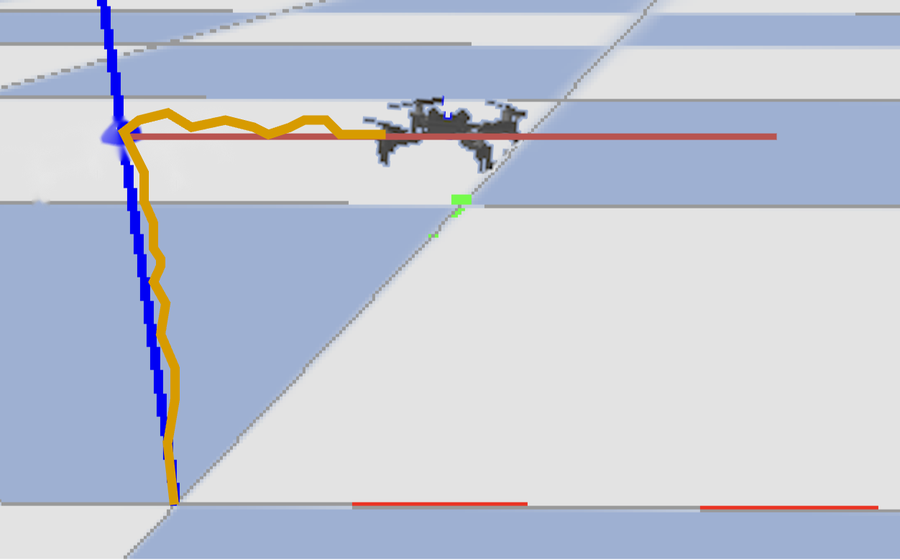} &
    \includegraphics[width=0.24\linewidth]{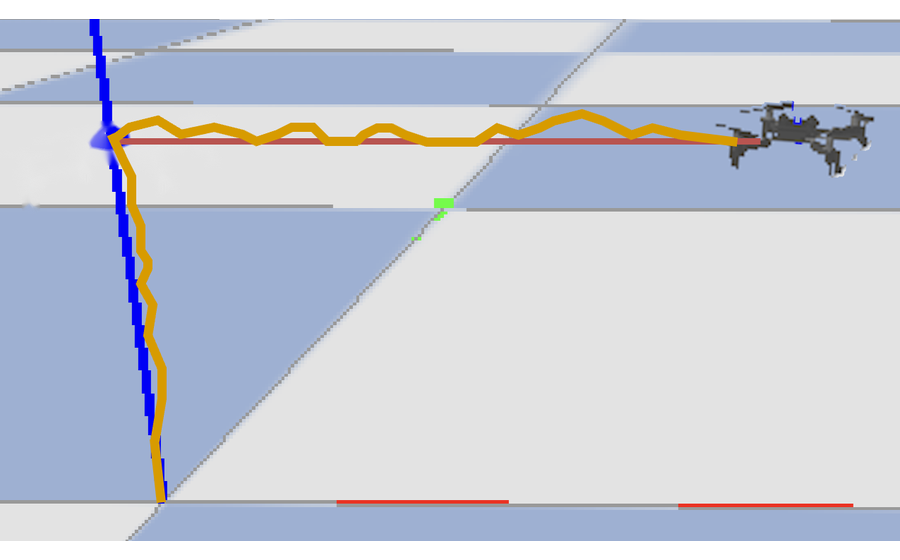} \\
    
    \end{tabular}
    \caption{Flight trajectories under \emph{all} action attack. 
    \textbf{Top:} Baseline-RL, \textbf{Middle:} ARMOR, and \textbf{Bottom:} \technique{}, which keeps its intended trajectory despite the attack.}
    \label{fig:trajectory_grid}
    \end{figure}

To visualize \technique{}'s actions against one of the \attack{} discussed in Table~\ref{tab:attack-types},
we run the \emph{All} action attack on a mission that follows the blue line and then turns right along the red line,
shown in Figure~\ref{fig:trajectory_grid} together with each technique's resulting trajectory in orange,
for all three techniques: baseline-RL (top), ARMOR (middle), and \technique{} (bottom). We can see that 
both baseline-RL and ARMOR cause the UAV to become unstable and crash, failing to finish the mission,
while \technique{} maintains the UAV's flight and stays on course until the mission completes. 

\subsection{Comparison with ARMOR and Baseline-RL}
\label{subsec:comparison}

%
We compare \technique{} against ARMOR and baseline-RL controller on the five \attack{} configurations from Table~\ref{tab:attack-types}, with results shown in Table~\ref{tab:physical_attack}.
Across the four single-channel attacks, compared to the baseline-RL controller, ARMOR increases the  average success rate from 40.5\% to around 50\% and reduces the average crash rate from 47\% to 36.5\%, with average state drifts of $\sim$0.41m and $\sim$0.39m, respectively.
Under the \emph{All} attack, however, both techniques cause  every mission to fail (0\% success, 100\% crash), with state drifts of $\sim$0.36m and $\sim$0.37m.
In contrast, \technique{} achieves an average of 95\% success with no crashes across the four single-channel attacks, and 67\% success with a 10\% crash rate under the \emph{All} attack, with the lowest average state drift of $\sim$0.30m and $\sim$0.26m, respectively, for the single channel and all attack.

\textbf{Takeaway:} These results show that using only RL (i.e., the baseline-RL controller), or its refinement with a two-phase teacher--student pipeline as in ARMOR, is insufficient against \attack{} and leaves the UAV vulnerable to action-space attacks.
In contrast, \technique{} makes the UAV resilient to action-space attacks, across all five configurations.

\begin{table*}[ht]
    \centering
    \caption{Performance comparison of Baseline-RL, ARMOR, and \technique{} under action attacks against five UAV action targets.}
    \label{tab:physical_attack}
    \renewcommand{\arraystretch}{1.15}
    \begin{tabular}{c|ccc|ccc|ccc}
    \hline
    \multirow{2}{*}{\textbf{Target action}} &
    \multicolumn{3}{c|}{\textbf{Baseline-RL}} &
    \multicolumn{3}{c|}{\textbf{ARMOR}} &
    \multicolumn{3}{c}{\textbf{\technique{}}} \\
    \cline{2-10}
    & \textbf{Success} & \textbf{Crash} & \textbf{State Drift (m)}
    & \textbf{Success} & \textbf{Crash} & \textbf{State Drift (m)}
    & \textbf{Success} & \textbf{Crash} & \textbf{State Drift (m)} \\
    \hline
    Pitch           & 18\% & 69\% & 0.307 $\pm$ 0.012 & 25\% & 57\%  & 0.287 $\pm$ 0.008 & 95\% & 0\% & 0.159 $\pm$ 0.004 \\
    Roll     & 19\% & 80\% & 0.418 $\pm$ 0.003 & 23\% & 75\%  &  0.428 $\pm$ 0.010 & 94\% & 0\% & 0.393 $\pm$ 0.003 \\
    Thrust & 61\% & 3\% & 0.608 $\pm$ 0.010 & 68\% & 0\% & 0.567 $\pm$ 0.034  & 97\% & 0\% & 0.394 $\pm$ 0.001 \\
    Gain  & 64\% & 35\% & 0.304 $\pm$ 0.001 & 85\% & 14\% & 0.295 $\pm$ 0.001  & 94\% & 0\% & 0.267 $\pm$ 0.001 \\
    All  & 0\% & 100\% & 0.359 $\pm$ 0.001 & 0\% & 100\% & 0.368 $\pm$ 0.001 & 67\% & 10\% & 0.259 $\pm$ 0.001 \\
    \hline
    \end{tabular}
    \end{table*}

\begin{figure}[h]
    \centering
    \includegraphics[width=\columnwidth]{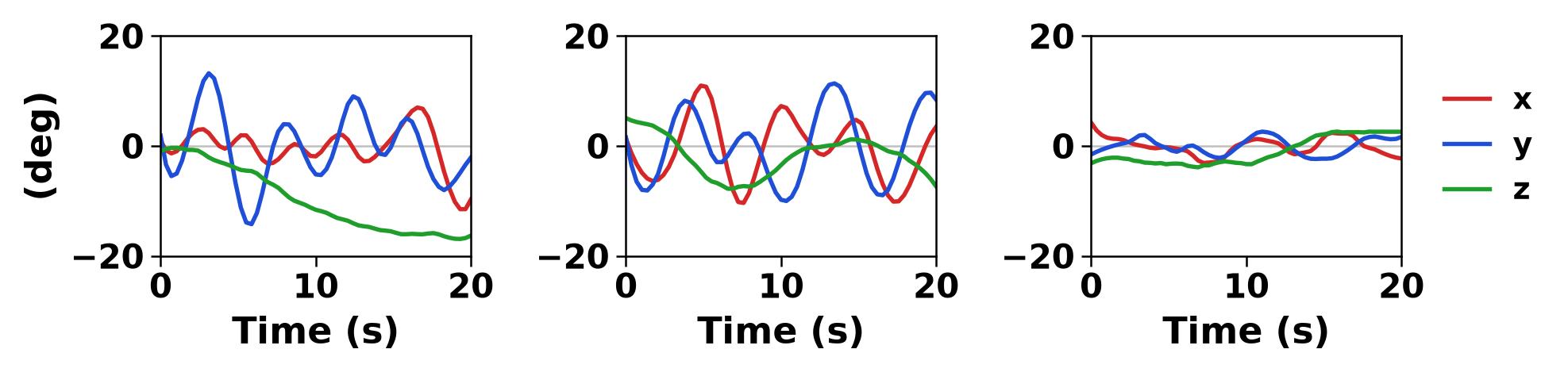}
    \caption{Attitude errors (degree) under Thrust action attack for Baseline-RL (left), ARMOR (middle), and \technique{} (right).}
    \label{fig:combined_attitude_error}
\end{figure}

Figure~\ref{fig:combined_attitude_error} illustrates this trend visually, showing the attitude error of the three techniques under the Thrust attack.
Baseline-RL's error exceeds $\pm 10^\circ$ on average, and ARMOR only marginally reduces it to around $\pm 10^\circ$, 
with both exhibiting large fluctuations and errors that cause crashes, consistent with the results in Table~\ref{tab:physical_attack}.
\technique{}, in contrast, keeps the average attitude error below $\pm 4^\circ$ throughout the attack, demonstrating its resilience against \attack{} such as the Thrust attack.



\subsection{\technique{} Zero-Shot Performance}
\label{subsec:zeroshot}

We evaluated the generalization of \technique{} by testing whether it remains resilient against \attack{} on action channels that were not part of its training data.
We consider two zero-shot scenarios: in each, we train \technique{} and ARMOR on a single randomly picked action attack (Pitch in one scenario, Gain in the other) and evaluate them on the attacks on the remaining three action channels from Table~\ref{tab:attack-types}.

\textbf{Pitch-trained.} We train \technique{} and ARMOR using Pitch as the training attack and evaluate on Roll, Thrust, and Gain, with the results in Table~\ref{tab:zero-shot-performance}.
ARMOR finishes an average of 49\% of missions with a 35\% crash rate, while \technique{} finishes 77\% of missions with only 7\% crashes and a lower average state drift ($\sim$0.33m vs. $\sim$0.44m for ARMOR).

\begin{table}[H]
    \centering
    \caption{\technique{} and ARMOR effectiveness when trained on \textbf{Pitch} action attacks only and tested on unseen attacks (\textit{Zero-shot}).}
    \label{tab:zero-shot-performance}
    \resizebox{\columnwidth}{!}{%
    \begin{tabular}{|c|c|c|c|c|}
    \hline
    \textbf{Techniques} & \textbf{Metrics} & \textbf{Roll} & \textbf{Thrust} & \textbf{Gain}\\
    \hline
    \multirow{3}{*}{ARMOR} & Success & 18\% & 50\% & 80\%\\
    & Crash & 81\% & 10\% & 15\%\\
    & State Drift (m) & 0.430 $\pm$ 0.014 & 0.597 $\pm$ 0.036 & 0.295 $\pm$ 0.001\\
    \hline
    \multirow{3}{*}{\technique{}} & Success & 80\% & 60\% & 91\%\\
    & Crash & 12\% & 6\% & 5\% \\
    & State Drift (m) & 0.375 $\pm$ 0.063 & 0.356 $\pm$ 0.072 & 0.270 $\pm$ 0.016\\
    \hline
    \end{tabular}%
    }
    \end{table}

\textbf{Gain-trained.} We use Gain as the training attack and evaluate on Pitch, Roll, and Thrust, with the results in Table~\ref{tab:zero-shot-performance2}.
ARMOR finishes 31\% of missions with a 57\% crash rate, whereas \technique{} finishes around 74\% of missions with only 9\% crashes and a lower average state drift ($\sim$0.34m vs. $\sim$0.44m for ARMOR).

\begin{table}[H]
    \centering
    \caption{\technique{} and ARMOR effectiveness when trained on \textbf{Gain} action attacks only and tested on unseen attacks (\textit{Zero-shot}).}
    \label{tab:zero-shot-performance2}
    \resizebox{\columnwidth}{!}{%
    \begin{tabular}{|c|c|c|c|c|}
    \hline
    \textbf{Techniques} & \textbf{Metrics} & \textbf{Pitch} & \textbf{Roll} & \textbf{Thrust}  \\
    \hline
    \multirow{3}{*}{ARMOR} & Success & 16\% & 14\% & 64\%\\
    & Crash & 82\% & 86\% & 4\%\\
    & State Drift (m) & 0.315 $\pm$ 0.009 & 0.436 $\pm$ 0.008 & 0.579 $\pm$ 0.050\\
    \hline
    \multirow{3}{*}{\technique{}} & Success & 44\% & 81\% & 96\% \\
    & Crash & 10\% & 18\% & 0\% \\
    & State Drift (m) & 0.231  $\pm$ 0.006 & 0.385 $\pm$ 0.002 & 0.395 $\pm$ 0.000\\
    \hline
    \end{tabular}%
    }
    \end{table}
    
\textbf{Takeaway:} Training on a single action attack is sufficient for \technique{} to generalize to unseen attack channels, achieving roughly $2$--$3\times$ ARMOR's success rate and a $5$--$6\times$ lower crash rate.


\subsection{\technique{} under Stealthy Attacks}
\label{subsec:stealthy}

To evaluate the effectiveness of \technique{} against stealthy attacks,
we adapt two attack profiles from prior work~\cite{dash2025armor} --- continuously increases the bias at a fixed rate,
and increases the bias in discrete increments at fixed intervals ---
to \attack{}, within the ranges discussed in Table~\ref{tab:attack-types}.
As can be seen in Table~\ref{tab:stealthy_attacks}, under these gradually accumulating attacks for Roll, 
\technique{} maintains a mission success rate of 100\% with no crashes, while ARMOR fails on every evaluated episode, 
resulting in a 0\% success rate and a 100\% crash rate. 

\textbf{Takeaway.} \technique{}'s pipeline encodes the history of actions into the student latent; the monitor reads this latent 
and continuously tracks the bias as it accumulates, adjusting its corrections as the disturbance grows and preventing large deviations.
In contrast, ARMOR lacks any runtime correction mechanism and fails regardless of how gradually the attack develops.
This suggests that the absence of an active correction stage, rather than the abruptness of the disturbance, 
is the primary factor behind ARMOR's vulnerability to \attack{}.

\begin{table}[H]
    \centering
    \caption{\technique{} and ARMOR performance under stealthy \attack{}.}
    \label{tab:stealthy_attacks}
    \begin{tabular}{|c|c|c|c|}
    \hline
    \textbf{Techniques} & \textbf{Success} & \textbf{Crash} & \textbf{State Drift (m)} \\
    \hline
    ARMOR & 0\% & 100\% & 0.419 $\pm$ 0.002 \\
    \hline
    \technique{} & 100\%  & 0\% & 0.392 $\pm$ 0.002 \\
    \hline
    \end{tabular}
    \end{table}

%% file: Sections/Conclusion.tex
\section{Discussion \& Conclusion}

RL controllers for UAVs are vulnerable to \attack{}, and while prior work addresses this by robustifying the policy through adversarial training, 
these defenses forward every action directly to the actuators once deployed — with no mechanism to intercept a command that has already been corrupted.
\technique{} closes this gap through three innovations: an \emph{action-attack-aware} latent trained with attack-specific privileged information, 
a two-phase teacher--student scheme that, as shown in Figure~\ref{fig:phoenix-training}(b), 
trains more efficiently and with fewer fluctuations than prior work while distilling this capability to a student that runs 
from physical-state and action history alone at deployment, and a monitor placed between the control policy and the actuators 
that uses this latent to intercept and correct corrupted commands before they reach the hardware.
Across attack scenarios where neither ARMOR nor an RL-only controller is resilient, \technique{} completes over 95\% of missions under single-channel attacks and 67\% under a combined attack on all four channels.
\technique{} further generalizes to attack channels unseen during training and maintains resilience against stealthy attack patterns.
However, extending \technique{} to other types of robotic systems and sim-to-real transfer remains an open direction for future work.